\documentclass[12pt]{article}

\usepackage[a4paper,left=20mm,right=20mm,top=25mm,bottom=30mm]{geometry}
\usepackage{amsfonts}
\usepackage{amsmath}
\usepackage{graphicx}
\usepackage{hyperref}
\usepackage{orcidlink}

\newenvironment{keywords}
{\par\noindent\textbf{Keywords: }}
{\par}

\title{A Low-Cost Hybrid Reservoir Computing Model for Isolated Sign Language Video Recognition}

\author{
Nitin Kumar Singh\textsuperscript{1,2,3,*}\,\orcidlink{0000-0001-6368-8437},
Arie Rachmad Syulistyo\textsuperscript{5}\,\orcidlink{0000-0002-5933-1168},
Yuichiro Tanaka\textsuperscript{3,4}\,\orcidlink{0000-0001-6974-070X},\\
Hakaru Tamukoh\textsuperscript{3,4}\,\orcidlink{0000-0002-3669-1371}
}

\date{
\textsuperscript{1}Woosong University Kazakhstan, Turkistan City, Kazakhstan\\
\textsuperscript{2}Data Science and Artificial Intelligence Laboratory (DSAIL), Woosong University Kazakhstan, Turkistan City, Kazakhstan\\
\textsuperscript{3}Graduate School of Life Science and Systems Engineering, Kyushu Institute of Technology, 2-4 Hibikino, Wakamatsu, Kitakyushu, 808-0196, Japan\\
\textsuperscript{4}Research Center for Neuromorphic AI Hardware, Kyushu Institute of Technology, 2-4 Hibikino, Wakamatsu, Kitakyushu, 808-0196, Japan\\
\textsuperscript{5}Department of Information Technology, State Polytechnic of Malang, Indonesia\\[1ex]
{\textsuperscript{*}{Corresponding author:} nitin@live.wsuk.edu.kz, nitinmjpruiitp@gmail.com}
}

\begin{document}

\maketitle

\begin{abstract}
Sign language recognition (SLR) enhances communication between hearing and hearing-impaired individuals. Although deep learning (DL) has achieved promising performance in SLR, its high computational cost limits deployment on edge devices. To address this challenge, we propose a lightweight reservoir computing (RC)-based approach for SLR. In the proposed method, MediaPipe extracts body and hand keypoints to capture the spatial and temporal dynamics of gestures. These keypoints are then processed by a hybrid reservoir computing (HRC) architecture that combines deep reservoir computing (DRC) and bidirectional reservoir computing (BRC), transforming the input into a high-dimensional dynamic representation. A ridge regression model maps the final HRC state to class labels. This HRC-based SLR method achieved Top-1, Top-5, and Top-10 accuracies of {61.12\%}, {86.05\%}, and {92.56\%}, respectively, on the Word-Level American Sign Language 100 (WLASL100) video dataset, demonstrating competitive performance compared to deep learning-based approaches. Additionally, due to the lightweight nature of RC, the training time was drastically reduced to only a few seconds compared with DL-based methods such as Bi-GRU.
This method offers low computational cost, showing its potential for deployment on edge devices.

\end{abstract}


\section{Introduction}
SLR is a vital area of research that aims to bridge the communication gap between deaf and hearing-impaired people. Unlike spoken language, sign language involves visual-spatial elements, where hand gestures, body movements, and facial expressions convey meaning~\cite{1}. SLR focuses on understanding human communication through visual motion patterns, including hand gestures and body movements. Unlike spoken language, sign language relies heavily on temporal coordination and spatial articulation, making its automatic recognition a challenging sequence modeling problem. In the current scenario, the global demand for assistive communication technologies is continuously increasing; consequently, efficient sign language recognition (SLR) systems have become crucial~\cite{2}.

Recent progress in SLR has largely been driven by deep learning models, which can automatically learn complex representations from large-scale video data~\cite{3,4,5}. Despite their effectiveness, such models typically require extensive training, large parameter sets, and high computational resources~\cite{6,7,8,9}. These requirements limit their suitability for real-time applications and deployment on low-power or embedded platforms, motivating the exploration of alternative modelling paradigms.

Zheng et al.~\cite{10} provide a comprehensive review of deep learning-based SLR systems, highlighting challenges such as extensive training time and high computational requirements. Koller et al. present a hybrid CNN-HMM (hidden Markov models) framework for continuous SLR and note that the combined model demands significant computational power, underscoring the need for more efficient and scalable approaches~\cite{11}.

To alleviate these issues, several model compression and optimization strategies, such as quantization, pruning, and lightweight network design, have been proposed to reduce the computational cost of deep learning models~\cite{12}. Although these techniques improve deployment feasibility to some extent, they often introduce additional training complexity, accuracy degradation, and hardware-dependent optimization overhead. Moreover, compressed deep learning models still rely on multi-layer backpropagation-based training and large parameter spaces, which limit their scalability and adaptability in real-time, low-power SLR applications, especially on edge devices~\cite{13}.

Collectively, these studies reveal that deep learning-based SLR systems face practical limitations, making them unsuitable for edge devices due to their extensive training requirements and computational overhead. As a result of these drawbacks, researchers are exploring alternative methods for sign language recognition.

One promising alternative to the above-mentioned challenges is reservoir computing, which offers an efficient paradigm for sequential data processing. RC models, such as echo state networks (ESNs), employ fixed internal reservoir weights, and only a lightweight linear readout layer is trained, resulting in drastically reduced training time and computational cost compared to deep learning-based models~\cite{14}. In contrast to backpropagation-based deep learning, the reservoir weights in an ESN are not trained; consequently, only the output layer is learned, making RC fast and resource-efficient~\cite{15,16}.

This work aims to improve the performance and efficiency of SLR systems, particularly by reducing training time and making them feasible for deployment on resource-constrained devices. To achieve this, we propose a system that combines Hybrid Reservoir Computing (HRC) with MediaPipe-based feature extraction. MediaPipe works as a lightweight and efficient frontend that extracts high-quality hand and body landmark features from video input, reducing the dimensionality and complexity of the input features passed to the learning model~\cite{17}.

The hybrid reservoir configuration combines bidirectional reservoirs with cascaded serial reservoirs (two stacked reservoirs), forming a hierarchical (cascaded or deep) configuration. This configuration enables the model to process information at multiple levels of temporal abstraction. In the DRC configuration, the first reservoir captures low-level short-term dynamics in the gesture sequence, while the second reservoir, fed by the output of the first, extracts higher-level long-term dependencies. Despite its hierarchical structure, the training process remains lightweight, as only the readout layer is trained. Additionally, the use of bidirectional connections within the reservoirs allows the system to process input sequences in both forward and backward directions, enhancing its ability to recognize gestures that depend on future and past context.

Unlike traditional deep neural networks, our HRC approach trains only the output readout layer, using simple linear regression rather than full backpropagation. This drastically reduces the training time while still enabling rich temporal feature representation through the layered reservoir structure. As a result, the proposed architecture achieves competitive accuracy while maintaining drastically low training time, remaining computationally lightweight and well-suited for edge deployment.

\section{Material and methods}
In this article, we use the Word-Level American Sign Language (WLASL) 100 video dataset, MediaPipe for extracting key features from WLASL videos, and an RC-based architecture for the classification of the sign language video dataset, which is explained in detail below:

\subsection{Dataset description}
\begin{figure}[!htbp]
    \centering
    \includegraphics[width=\linewidth]{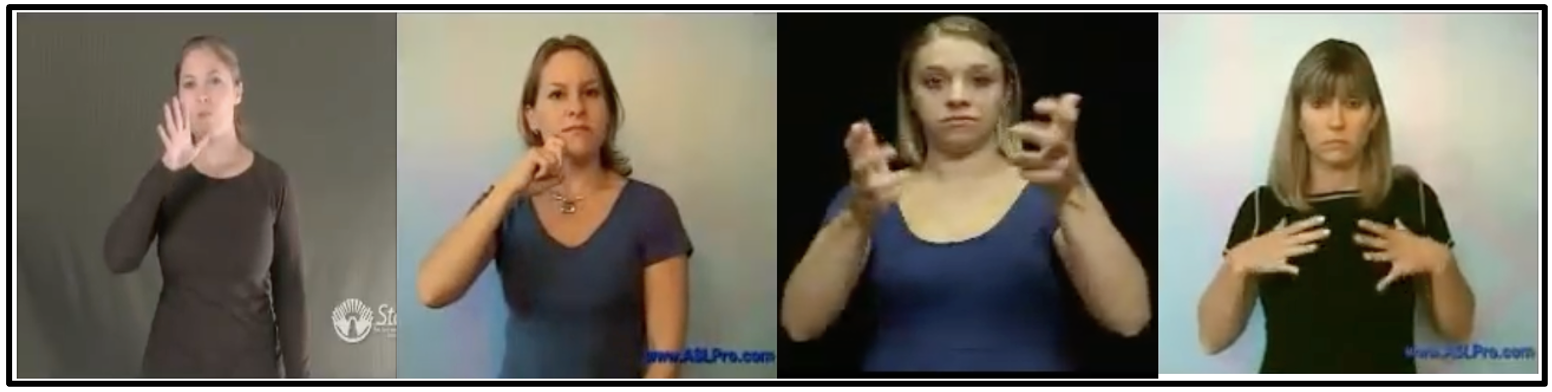}
    \caption{Signs used by different signers for various activities.}
    \label{fig_signs}
\end{figure}

This study utilized the WLASL100 sign language recognition video dataset~\cite{18}. WLASL100 is a 100-class subset of the larger WLASL dataset and is commonly used to evaluate isolated sign language recognition systems. The dataset contains videos performed by different signers, capturing natural variations in gesture appearance, movement speed, and signing style across the sign categories, as shown in Fig.~\ref{fig_signs}.

The WLASL100 dataset comprises 100 distinct sign classes and includes 1,780 videos for training, 258 videos for validation, and 258 videos for testing~\cite{18}.

\subsection{MediaPipe}

We used MediaPipe to extract key features from the WLASL100 video dataset.  
Google's MediaPipe is an open-source framework for extracting various features from video inputs, as shown in Fig.~\ref{fig_feature_extraction}. It is particularly useful for tasks such as SLR. MediaPipe provides hand and pose detection models, which are essential for SLR~\cite{19}. These models detect and track keypoints on the hands and body, capturing the spatial and temporal dynamics of sign language gestures. We used MediaPipe's hand-tracking and pose-detection features to extract key features from the WLASL100 video dataset, as explained briefly below:

\begin{figure}[!htbp]
    \centering
    \includegraphics[width=\linewidth]{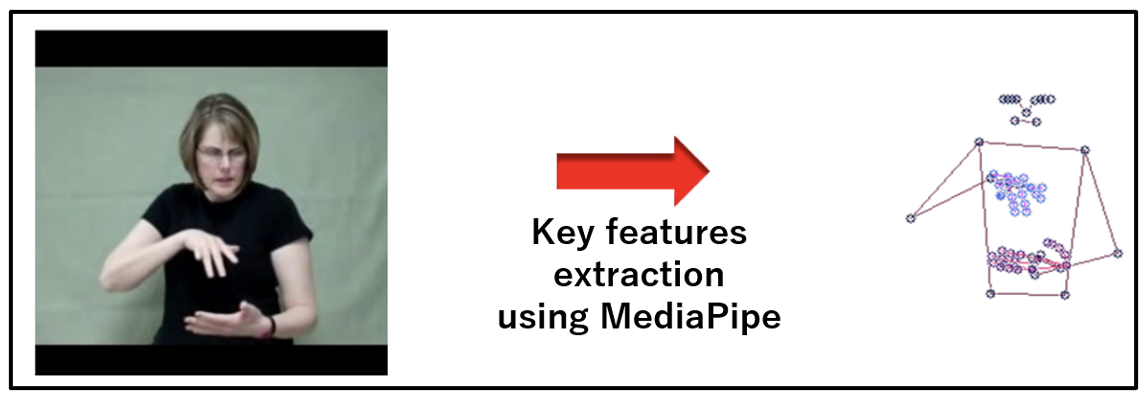}
    \caption{Feature extraction using MediaPipe.}
    \label{fig_feature_extraction}
\end{figure}

\begin{itemize}
    \item {Hand tracking:} MediaPipe's hand-tracking model identifies 21 keypoints on each hand, including the wrist, knuckles, and fingertips. This detailed mapping allows for precise tracking of hand movements and gestures.
    \item {Pose detection:} The pose-detection model identifies 33 keypoints on the human body, encompassing the head, shoulders, elbows, wrists, hips, knees, and ankles. This comprehensive body mapping aids in understanding the overall posture and movement dynamics during sign language communication.
\end{itemize}

As explained, MediaPipe provides a suite of solutions for processing video data, including hand tracking, pose estimation, and face detection models.

\subsection{Reservoir Computing}

RC consists of a fixed, sparsely, and randomly connected recurrent neural network, referred to as the reservoir. The internal weights of the reservoir remain untrained. When input signals are projected into this high-dimensional dynamical system, the reservoir's nonlinear response encodes temporal dependencies. Although the internal structure of the reservoir remains unchanged, it produces dynamic responses to inputs, generating high-dimensional representations that capture complex temporal patterns~\cite{20}. As a result, RC offers fast training and low computational cost, making it well suited for time-dependent tasks in which computational efficiency is important.

Unlike traditional deep learning-based models that require the entire architecture to undergo extensive training, RC focuses on training only the output layer. 
In this work, we evaluate different RC configurations, including a standard ESN with a single unidirectional reservoir, DRC, BRC, and the proposed HRC architecture.

\subsubsection{Echo State Networks (ESNs)}
\begin{figure}[!htbp]
    \centering
    \includegraphics[width=\linewidth]{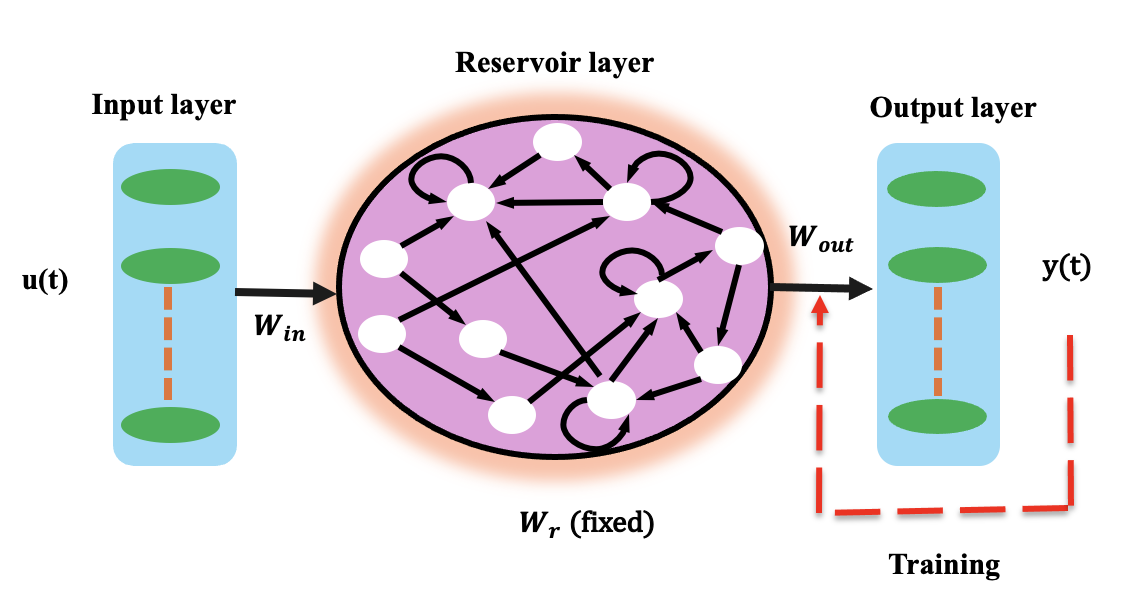}
    \caption{Echo State Network (ESN)}
    \label{fig_esn_rc}
\end{figure}

ESNs belong to the family of reservoir computing methods and can efficiently process temporal data. ESNs are governed by the echo state property, ensuring that the influence of past inputs gradually fades over time. The reservoir in an ESN is initialized with random and sparse recurrent connections and operates in a stable dynamic regime controlled by parameters such as the spectral radius. During operation, the reservoir generates a sequence of internal states driven by the input signal, while learning is limited to a linear readout that maps these states to the target output. This design allows ESNs to model temporal dependencies effectively while avoiding iterative weight updates inside the recurrent structure, allowing temporal patterns to be learned with low computational cost. A key feature of ESNs is that only the output layer is trained, while the internal connections within the reservoir remain unchanged, allowing for fast training and simplified optimization~\cite{21,22}. The schematic diagram of ESN-based RC is shown in Fig.~\ref{fig_esn_rc}.

Let $u(t) \in \mathbb{R}^{N_u}$ represent the input at time $t$, $x(t) \in \mathbb{R}^{N_r}$ be the reservoir state, and $y(t) \in \mathbb{R}^{N_y}$ be the output. The weight matrices include the input weight matrix $W_{\text{in}} \in \mathbb{R}^{N_r \times N_u}$, the recurrent reservoir weight matrix $W_r \in \mathbb{R}^{N_r \times N_r}$, and the output weight matrix $W_{\text{out}} \in \mathbb{R}^{N_y \times N_r}$. The scalar $\alpha \in [0,1]$ represents the leak rate, and $f(\cdot)$ is a nonlinear activation function such as $\tanh$.

The reservoir dynamics are governed by the state update equation, as shown in Equation (1):
\noindent
\begin{equation}
x(t) = (1 - \alpha)x(t-1) + \alpha f\left(W_{\text{in}}u(t) + W_r x(t-1)\right)
\end{equation}

The output is computed as a linear combination of the reservoir state, as shown in Equation (2):
\begin{equation}
y(t) = W_{\text{out}} x(t)
\end{equation}

During training, only the output weights $W_{\text{out}}$ are optimized. Given a matrix $X \in \mathbb{R}^{N_r \times T}$ of collected reservoir states over $T$ time steps, and the corresponding target outputs $Y \in \mathbb{R}^{N_y \times T}$, the readout weights can be computed using ridge regression, as shown in Equation (3):
\noindent
\begin{equation}
W_{\text{out}} = Y X^\top (X X^\top + \lambda I)^{-1}
\end{equation}

For the reservoir to exhibit meaningful dynamics, it must satisfy the Echo State Property (ESP). A sufficient condition for ESP is given by Equation (4):
\noindent
\begin{equation}
\rho(W_r) < 1
\end{equation}
where $\rho(W_r)$ denotes the spectral radius of the reservoir weight matrix. The spectral radius is typically set slightly below one to ensure the reservoir remains stable while maintaining rich dynamics~\cite{23}. The leak rate controls the memory depth of the reservoir, with smaller values allowing the network to capture longer temporal dependencies.

ESNs offer significant advantages for processing sequential data, including fast training and low computational cost, making them well-suited for deployment on edge devices with limited resources.

\subsubsection{Deep Reservoir Computing (DRC)}

Deep Reservoir Computing is an extension of traditional RC, in which multiple reservoirs are stacked in layers, enabling hierarchical processing of temporal features and enhanced representation of complex sequential data.
The DRC framework utilizes multiple reservoirs connected in a cascaded manner to model hierarchical temporal features in time-series data. Each reservoir processes inputs, with its outputs feeding into the next, enabling the system to capture both short-term and long-term dependencies. We used two ESN reservoirs connected in a cascaded configuration~\cite{24,25,26}. We used ridge regression to train only the output weights, thereby simplifying optimization and transforming the reservoir dynamics into compact features for classification~\cite{27}. 

\begin{figure}[!htbp]
    \centering
    \includegraphics[width=\linewidth]{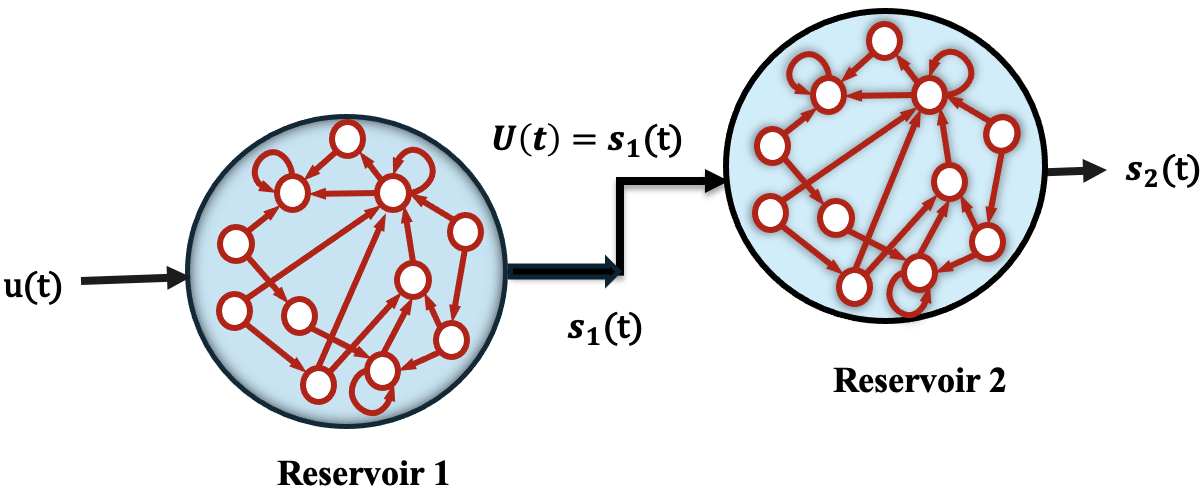}
    \caption{Deep reservoir computing}
    \label{fig_deep_rc}
\end{figure}

In a DRC framework, multiple reservoir layers are connected in a cascaded manner (hierarchically).
Each reservoir layer processes the temporal dynamics produced by the preceding layer, allowing hierarchical temporal representations to be formed.

Let $s_l(t)$ denote the state of the $l$-th reservoir layer $(l = 1, 2, \dots, N)$ at time $t$, where $N$ denotes the total number of reservoir layers in the model.
The state update equation for each reservoir layer is given by

\begin{equation}
\label{eq:deep_reservoir_general}
s_l(t) = (1 - \alpha_l)\, s_l(t-1) + \alpha_l\, \tanh\!\left( W_{r}^{(l)}\, s_l(t-1) + W_{\text{in}}^{(l)}\, x_l(t) \right)
\end{equation}

where $\alpha_l$ denotes the leak rate of the $l$-th reservoir,
$W_{r}^{(l)}$ is the recurrent weight matrix,
and $W_{\text{in}}^{(l)}$ is the input weight matrix of the $l$-th reservoir layer.
The input to each layer, $x_l(t)$, is defined as

\begin{equation}
x_l(t) =
\begin{cases}
u(t), & \text{if } l = 1, \\
s_{l-1}(t), & \text{if } l > 1
\end{cases}
\end{equation}

where $u(t)$ denotes the external input signal at time $t$.
In this study, the deep reservoir architecture consists of two reservoirs connected serially, forming a cascaded structure with $N = 2$.
In this configuration, each reservoir receives the state of the preceding layer, allowing temporal dependencies to be captured hierarchically.

\subsubsection{Bidirectional Reservoir Computing (BRC)}

\begin{figure}[!htbp]
    \centering
    \includegraphics[width=\linewidth]{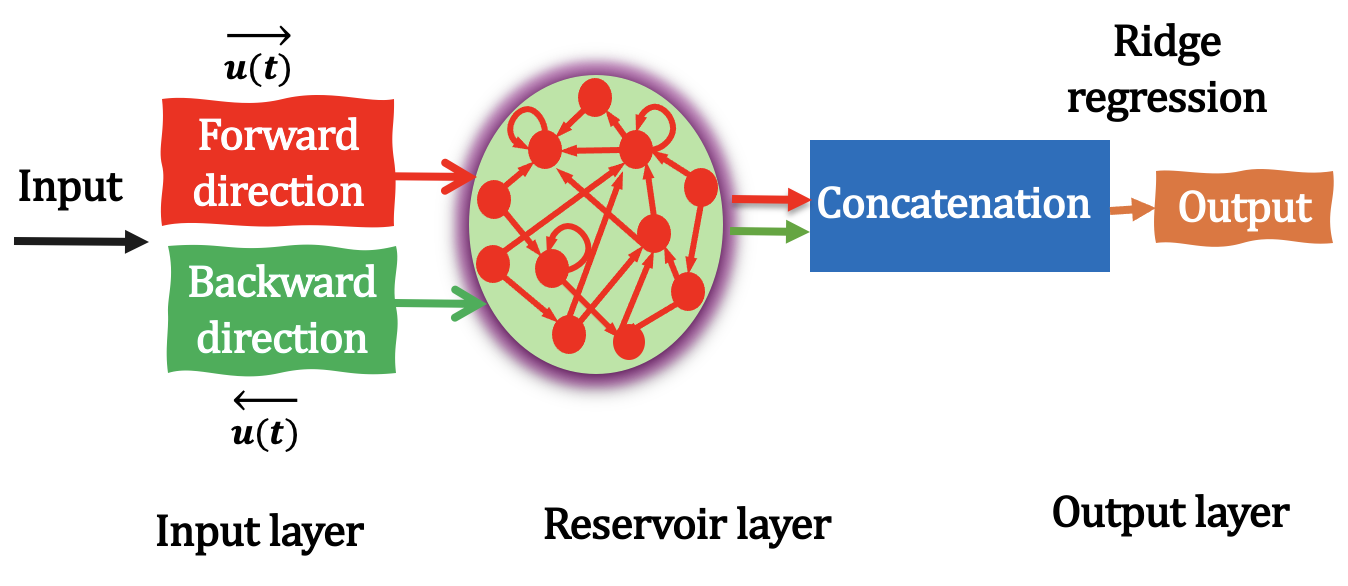}
    \caption{Bidirectioanl reservoir computing}
    \label{fig_bidirectional_rc}
\end{figure}

Bidirectional reservoir computing processes the input sequence in both forward and backward directions, enabling the model to capture information from both past and future contexts~\cite{28,29}. In a standard ESN, inputs flow in one direction, affecting the reservoir's state and influencing the output. However, a bidirectional ESN introduces a second pathway that processes the input sequence in reverse, enabling the model to capture information from both forward and backward temporal directions, as shown in Fig.~\ref{fig_bidirectional_rc}.

BRC enhances the model's ability to comprehend temporal sequences in applications such as SLR, speech, and text analysis, where both preceding and following contexts can influence the interpretation of the current state.

In a bidirectional ESN, the input sequence is processed in both forward and backward directions to generate high-dimensional reservoir states. These states are subsequently concatenated to perform prediction or classification by utilizing temporal information from the complete input sequence.

 The state update of the bidirectional reservoir can be expressed in a unified form as
\begin{equation}
x_d(t+1) = f\!\left( W x_d(t) + W_{\mathrm{in}}^{(d)} u_d(t) \right)
\label{eq:bidirectional_state_update}
\end{equation}

where $d \in \{f, b\}$ denotes the reservoir direction. For the forward reservoir ($d = f$), $u_f(t) = u(t)$ corresponds to the original input sequence, while for the backward reservoir ($d = b$), $u_b(t) = u(T - t)$ represents the time-reversed input sequence. Here, $x_d(t)$ is the reservoir state vector at time $t$, $W$ is the fixed recurrent weight matrix, $W_{\mathrm{in}}^{(d)}$ is the direction-specific input weight matrix, and $f(\cdot)$ denotes a nonlinear activation function. In this work, the hyperbolic tangent function is used.

The forward and backward reservoir states, corresponding to the original and time-reversed input sequences, are concatenated to form a unified bidirectional representation~\cite{30,31}. The final output is then computed as
\begin{equation}
y(t) = W_{\mathrm{out}} \big( x_f(t) \oplus x_b(t) \big)
\label{eq:combined_output}
\end{equation}
where $y(t)$ denotes the output vector at time $t$, $\oplus$ represents the concatenation operator, and $W_{\mathrm{out}}$ is the trainable readout weight matrix. The readout is trained using ridge regression, enabling efficient classification while preserving the complementary temporal information captured by the forward and backward reservoirs.

\subsubsection{Hybrid Reservoir Computing}

\begin{figure}[!htbp]
    \centering
    \includegraphics[width=\linewidth]{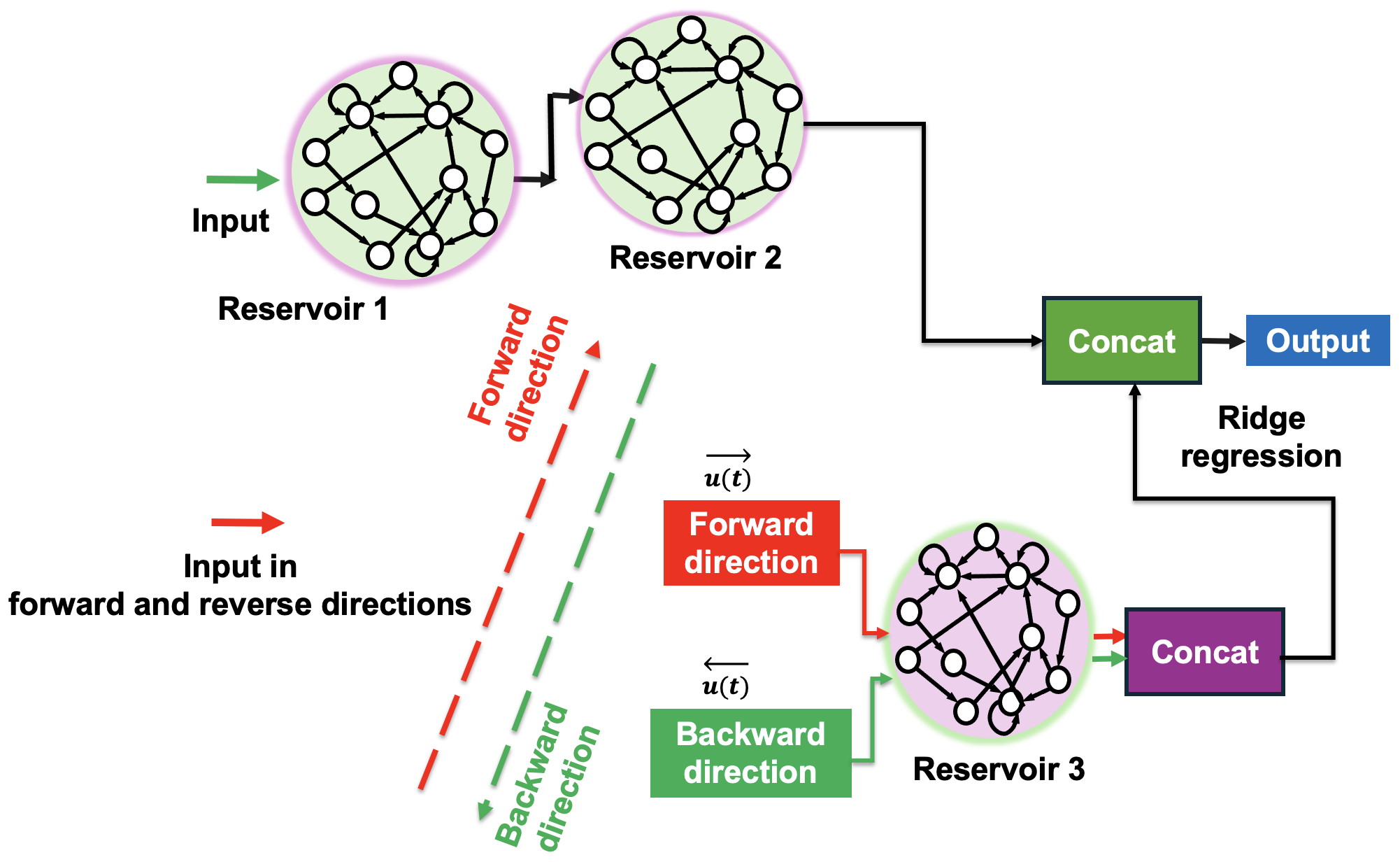}
    \caption{Hybrid reservoir computing.}
    \label{fig_hybrid_rc}
\end{figure}

Fig.~\ref{fig_hybrid_rc} illustrates an HRC-based architecture that integrates both DRC (two reservoirs cascaded in series) and BRC to model complex temporal dynamics more effectively. The Hybrid Reservoir Computing (HRC) architecture consists of a set of
reservoir units operating in bidirectional and cascaded configurations.
In the DRC unit, the input flows through two serially connected reservoirs (Reservoir~1 and Reservoir~2), allowing for hierarchical temporal abstraction, as each layer captures increasingly complex patterns over time. Simultaneously, in the BRC unit, the same input is processed in both the forward and backward directions through Reservoir~3, allowing the system to capture context from both past and future frames. The outputs from the forward and backward directions are concatenated to form a bidirectional temporal representation, as explained below.

The final HRC state is obtained by concatenating the states of all
reservoir units as
\begin{equation}
x_{\mathrm{HRC}}(t) = \bigoplus_{k=1}^{K} x_k(t)
\label{eq:hrc_fusion}
\end{equation}
where $\oplus$ denotes vector concatenation and $K$ is the total number
of reservoir units. The output is computed using a linear readout:
\begin{equation}
y(t) = W_{\mathrm{out}} x_{\mathrm{HRC}}(t)
\label{eq:hrc_readout_general}
\end{equation}

Finally, the outputs from both the deep reservoir path and the bidirectional reservoir path are combined via concatenation and passed through a ridge regression layer to generate the final prediction. This architecture combines hierarchical and bidirectional reservoir dynamics to produce richer temporal representations.

In this study, the HRC configuration is designed to process sequential data by leveraging a combination of deep and bidirectional reservoirs within an ESN framework.

\subsection{Experimental settings}
Here, we employed an ESN-based HRC for SLR, where each reservoir was configured according to the hybrid architecture. The HRC architecture consisted of two cascaded, serially connected ESN reservoirs (deep reservoir component), with each reservoir consisting of 100 nodes. In parallel, the ESN-based bidirectional reservoir component consisted of 120 nodes. We also compared our results with BRC (170 nodes) and DRC (170 nodes in each reservoir), standard ESN (340 nodes) as shown in Table~\ref{table:architecture_summary}. 

\begin{table}[ht]
\centering
\caption{Comparison of reservoir architectures used in this study}
\label{table:architecture_summary}
\renewcommand{\arraystretch}{1.0}
\setlength{\tabcolsep}{3pt}
\scriptsize
\begin{tabular}{|l|c|c|c|}
\hline
Method & No. of reservoirs & Total concatenated nodes & Architecture type \\
\hline
HRC (proposed) 
& 3 
& 340 
& Deep + Bidirectional \\

BRC 
& 1 
& 170 
& Bidirectional \\

DRC 
& 2 
& 340    
& Deep (serial) \\

ESN
& 1
& 340    
& Standard \\
\hline
\end{tabular}
\end{table}

MediaPipe was used to extract keypoints from the WLASL100 dataset in all methods. All experiments were performed on an Intel\textsuperscript{\textregistered} Core\textsuperscript{™} i7-11700 processor with 32~GB of RAM.

In this study, the spectral radius ($\rho$) was set to 0.1 and the leak rate ($\alpha$) to 0.9 for all reservoir configurations. Optuna can be used to optimize these parameters.

\subsection{Performance evaluation}
To assess the effectiveness of the proposed method, accuracy was calculated using the following equation~\cite{32}:
\noindent
\begin{equation}
    \text{Accuracy} = \frac{\text{True Positives} + \text{True Negatives}}{\text{Total Samples}}
\end{equation}

\section{Results and discussion}
\begin{figure}[!htbp]
    \centering
    \includegraphics[width=\linewidth]{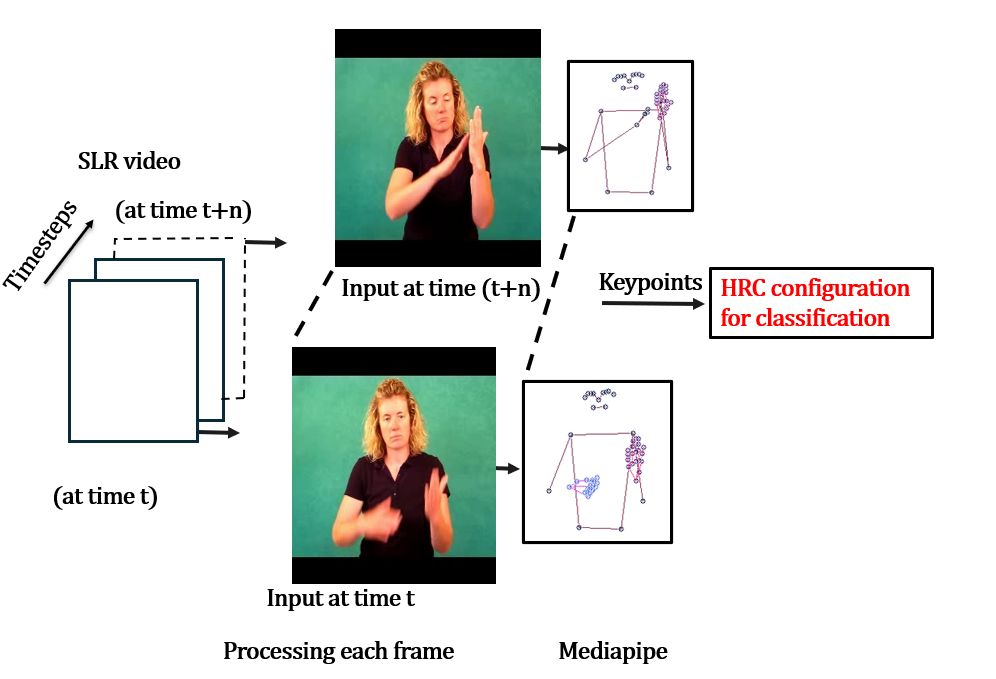}
    \caption{SLR system using hybrid reservoir computing.}
    \label{fig_slr_hybrid_rc}
\end{figure}

We compared the HRC framework against standard reservoir computing (ESN), DRC, BRC, and deep learning-based models for SLR to assess the competitiveness of the proposed method.
Fig.~\ref{fig_slr_hybrid_rc} illustrates the overall SLR framework using hybrid reservoir computing. As described in detail earlier, sign language videos are processed frame-wise using MediaPipe to extract skeletal landmarks, which are subsequently fed into the ESN-based HRC architecture for temporal modeling and classification.

As illustrated in Fig.~\ref{fig_slr_hybrid_rc}, the extracted keypoint features are processed by the hybrid reservoir computing framework, which combines deep and bidirectional reservoirs to model hierarchical and contextual temporal information. The resulting reservoir states are integrated and mapped to class labels using a ridge regression-based readout.

Table~\ref{table:reservoir_summary} compares the performance of the SLR system using HRC, BRC, DRC, Bi-GRU, and standard ESN on the WLASL100 dataset in terms of accuracy and training time. To ensure robustness, each experiment was conducted five times, and the mean accuracy was reported. In all cases, the error was less than 10\%.

\begin{table}[ht]
\centering
\caption{Comparison of SLR using HRC, BRC, DRC, Bi-GRU, and standard ESN on WLASL100}
\label{table:reservoir_summary}
\renewcommand{\arraystretch}{1.0}
\setlength{\tabcolsep}{3pt}
\scriptsize
\begin{tabular}{|l|c|c|c|}
\hline
Methods & Accuracy (\%) & Training time (mm:ss.ms) & Reference \\
\hline
HRC (proposed)  & 61.12 & 00:14.61 & {Proposed} \\ 
BRC             & 59.21 & 00:13.62 & {This work} \\ 
DRC             & 57.21 & 00:18.52 & {This work} \\ 
Standard ESN    & 56.29 & 00:13.17& {This work}  \\
Bi-GRU          & 50.36 & 33:54.1  & \cite{33} \\
\hline
\end{tabular}
\end{table}

From Table~\ref{table:reservoir_summary}, it can be concluded that the HRC model outperforms the standard ESN, BRC, DRC, and Bi-GRU, achieving the highest accuracy. It also demonstrates competitive training efficiency, requiring only 14.61 seconds on the CPU, which is significantly faster than the deep learning-based Bi-GRU baseline that takes approximately 33 minutes and 54 seconds on a CPU.

\begin{table}[ht]
\centering
\caption{Comparison of HRC for SLR on the WLASL100 dataset including DL-based approaches such as Pose-TGCN and I3D}
\label{tab:accuracy_comparison}
\renewcommand{\arraystretch}{1.0}
\setlength{\tabcolsep}{2.5pt}
\scriptsize
\begin{tabular}{|l|c|c|c|c|c|c|}
\hline
Method & \multicolumn{3}{c|}{Accuracy (\%)} & Inference time & Training time & Reference \\
\cline{2-4}
 & Top-1 & Top-5 & Top-10 & (mm:ss.ms) & (hh:mm:ss.ms) & \\
\hline
HRC       & 61.12 & 86.05 & 92.56 & 00:00:1.47  & 00:00:14.61 & This work \\
Pose-TGCN & 55.43 & 78.68 & 87.60 & 00:00:4.20  & 00:38:18.90 & \cite{33} \\
I3D       & 65.89 & 84.11 & 89.92 & 00:00:12.50 & 20:13:42.50 & \cite{33} \\
MRC       & 60.35 & 84.65 & 91.51 & 00:00:5.20  & 00:00:52.70 & \cite{33} \\
\hline
\end{tabular}
\end{table}

Table~\ref{tab:accuracy_comparison} further compares the Top-1, Top-5, and Top-10 accuracies of the proposed HRC-based architecture with deep learning approaches for SLR on the WLASL100 dataset. The proposed approach demonstrates competitive performance, achieving 61.12\% (Top-1), 86.05\% (Top-5), and 92.56\% (Top-10) accuracy, while requiring an inference time of 1.47 seconds.

Hosain \textit{et al.} also used a 3D convolutional neural network (3D CNN) for isolated American Sign Language recognition using the GMU-ASL51 dataset. The 3D CNN achieved an average accuracy of 52\% with a standard deviation of 12\%, which is lower than that of the proposed HRC. Additionally, training the 3D CNN is computationally expensive, requiring approximately 20 hours on an NVIDIA Tesla K80 GPU, highlighting the limitations of using 3D CNNs alone for efficient sign language recognition~\cite{34}.

Naz \textit{et al.} evaluated appearance-based models for isolated sign language recognition on the WLASL100 dataset. Based on the reported results, the I3D model achieved a top-10 accuracy of 89.92\%, while the temporal graph convolution network (TGCN) obtained a slightly lower top-10 accuracy of 87.60\%. These results indicate that spatiotemporal video-based architectures such as I3D provide competitive performance for isolated SLR on WLASL100, although they rely on deep architectures with higher computational complexity~\cite{35}.

Based on the results, the proposed method achieves competitive performance compared with deep learning-based SLR systems while maintaining substantially lower computational cost. Unlike deep models that require extensive parameter optimization, in RC, recurrent dynamics are generated by fixed, randomly connected reservoirs, allowing the model to explore a high-dimensional state space without iterative weight adaptation. Consequently, learning is restricted to a convex optimization problem at the readout stage, which avoids issues such as gradient instability and slow convergence. This property enables the preservation of rich temporal dynamics while ensuring stable and predictable training behavior, particularly when processing long or variable-length gesture sequences. Further, the hybrid architecture, which is a combination of cascaded (DRC) and bidirectional (BRC) configured reservoirs, enables the model to capture both short- and long-term gesture dynamics as well as contextual information from past and future frames. This combination allows HRC to achieve high accuracy with low latency, making it suited for sign language recognition in resource-constrained or edge-device environments.

\section{Conclusions}
This study investigated the use of MediaPipe and HRC for isolated SLR. MediaPipe was used to extract keypoints from the WLASL100 video dataset, which were subsequently processed by the HRC architecture for gesture classification. The proposed model achieved Top-1, Top-5, and Top-10 accuracies of 61.12\%, 86.05\%, and 92.56\%, respectively. By combining deep and bidirectional reservoir configurations, HRC captures hierarchical temporal patterns and contextual information from both directions, contributing to its competitive recognition performance. At the same time, the computational efficiency of the framework is inherited from the lightweight nature of the RC paradigm, in which the reservoir weights remain fixed and only the readout layer is trained. Consequently, the proposed method requires substantially less training time than the deep learning-based approaches used for SLR.
TThe results obtained in this study show that the HRC-based SLR system has the potential to be deployed on resource-constrained devices.
Future work may extend the framework to continuous SLR by incorporating temporal segmentation methods and improving the reservoir dynamics to handle longer and more complex gesture sequences.

\section*{Data Availability}

The dataset used in this work is available at:
\url{https://dxli94.github.io/WLASL/}.

\section*{Funding}

This work was financially supported in part by the New Energy and Industrial Technology Development Organization (NEDO), the Japan Science and Technology Agency (JST ALCA-Next), and the Japan Society for the Promotion of Science (JSPS KAKENHI). The corresponding grant numbers are JPNP16007 (NEDO), JPMJAN23F3 (JST ALCA-Next), and 22K17968, 23H03468, and 23K18495 (JSPS KAKENHI).

\section*{Conflict of Interest}

The authors declare that they have no conflict of interest.

\section*{Author Contributions}

Author~1 contributed to conceptualization, investigation, visualization, and writing the original draft. Authors~3 and~4 acquired funding and supervised the research. Authors~1, 2, 3, and~4 reviewed and edited the manuscript.

\section*{Appendix: Links}

\begin{tabular}{p{12cm}}
\url{https://doi.org/10.32545/ev202604001751.v1}\\[0.3cm]
\url{https://doi.org/10.32545/ev202604001394.v1}\\[0.3cm]
\url{https://www.youtube.com/@Nitin18632}
\end{tabular}

\end{document}